\pdfoutput=1

\documentclass[11pt]{article}

\usepackage[preprint]{acl}

\usepackage{times}
\usepackage{latexsym}
\usepackage{booktabs}
\usepackage{amsmath} 
\usepackage{amssymb}
\usepackage[T1]{fontenc}

\usepackage[utf8]{inputenc}

\usepackage{microtype}

\usepackage{inconsolata}

\usepackage{graphicx}

\title{R-LoRA: Randomized Multi-Head LoRA for Efficient Multi-Task Learning}

\author{
        Jinda Liu$^{1}$ \quad  Yi Chang$^{1,2,3}$ \quad Yuan Wu$^{1}$\footnotemark[1] \\
        $^{1}$School of Artificial Intelligence, Jilin University \\
        $^{2}$Engineering Research Center of Knowledge-Driven Human-Machine Intelligence, MOE, China \\
        $^{3}$International Center of Future Science, Jilin University\\
        liujd9922@mails.jlu.edu.cn, yichang@jlu.edu.cn, yuanwu@jlu.edu.cn \\
}

\begin{document}
\maketitle


\begin{abstract}
Fine-tuning large language models (LLMs) is computationally expensive, and Low-Rank Adaptation (LoRA) provides a cost-effective solution by approximating weight updates through low-rank matrices. In real-world scenarios, LLMs are fine-tuned on data from multiple domains to perform tasks across various fields, embodying multi-task learning (MTL). LoRA often underperforms in such complex scenarios. To enhance LoRA's capability in multi-task learning, we propose R-LoRA, which incorporates Multi-Head Randomization. Multi-Head Randomization diversifies the head matrices through Multi-Head Dropout and Multi-Head Random Initialization, enabling more efficient learning of task-specific features while maintaining shared knowledge representation. Our approach not only improves performance in MTL but also reduces GPU memory usage and training time. Experiments show that R-LoRA's gains stem from increased diversity in the head matrices, demonstrating its effectiveness for multi-task learning. The code is available at \url{https://github.com/jinda-liu/R-LoRA}
\end{abstract}

\begin{figure*}[t]
\centering
  \includegraphics[width=0.8\linewidth]{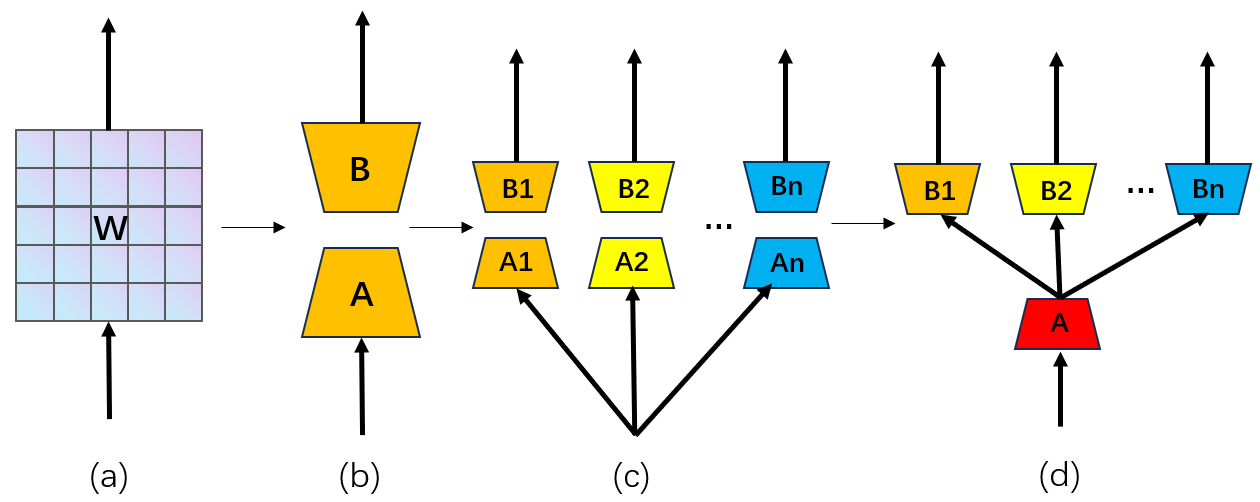}
  \caption {Training architecture comparison. (a) Full parameter fine-tuning; (b) Vanilla LoRA; (c) Multi-Adapter architecture; (d) Multi-Head/Asymmetric architecture.}
  \label{architecture}
\end{figure*}

\section{Introduction}

In recent years, large language models (LLMs) have manifested unprecedentedly superior performance in various natural language processing (NLP) tasks \cite{brown2020language,zhao2023survey,chang2024survey}. Due to its impressive capabilities in language understanding and generation, LLMs have gained extensive interest from both academia and industry. Despite their high generalizability, LLMs still require fine-tuning for specific domains or updating the knowledge base~\cite{agiza2024mtlora,xin2024beyond}. 

Supervised fine-tuning (SFT) is crucial for aligning large language models (LLMs) with human instructions, which trains the model with a small yet high-quality set of labeled data~\cite{hu2021lora,xia2024rethinking}. The vast number of parameters in LLMs poses significant challenges regarding computational efficiency and memory consumption during full fine-tuning (FT), which updates all parameters. 

To address the issue of hardware requirements for LLM adaptation, a solution called parameter efficient fine-tuning (PEFT) has been proposed~\cite{han2024parameter}. PEFT methods reduce VRAM usage of cached optimizer states by only optimizing a fraction of model parameters while keeping the rest frozen. Various PEFT methods have been widely studied\cite{li2021prefix}\cite{liu2024gpt}\cite{liu2022few}\cite{hu2021lora}. Among these methods, LoRA has emerged as the mainstream alternative to full parameter fine-tuning. Instead of updating the original parameter matrix directly, LoRA approximates the updated parameters using the product of two smaller matrices. During inference, the output obtained from the original parameter matrix is combined with the output from the updated parameter matrices. However, LoRA does not perform well in multi-task scenarios, particularly in dealing with complex datasets.

Recent LoRA variants have improved multi-task learning by employing multiple LoRA adapters, including Multi-LoRA~\cite{wang2023multilora}, LoRA-MoE~\cite{dou2023loramoe}, and MoeLoRA~\cite{liu2024moemeetsllmsparameter}. We refer to this extended framework as the Multi-Adapter LoRA architecture, which consists of multiple down-projection matrices (A) and their corresponding head matrices (B), enabling task-specific adaptation through diverse parameter sets. Notably, LoRA-MoE and MoeLoRA further enhance this architecture by introducing a Mixture of Experts (MoE) mechanism to aggregate adapter outputs.
\citet{tian2024hydraloraasymmetricloraarchitecture} observes that in the Multi-Adapter LoRA architecture, the parameters of the down-projection matrices A are relatively consistent, while the differences between the head matrices B are more pronounced, which aids in capturing task-specific knowledge. To leverage this property, HydraLoRA~\cite{tian2024hydraloraasymmetricloraarchitecture} is proposed to feature an asymmetric architecture with one shared down-projection matrix A and multiple task-specific head matrices B. Additionally, HydraLoRA also employs an MoE mechanism to aggregate the outputs of the head matrices. This design achieves a good balance between training performance and parameter efficiency. In this work, we explicitly define asymmetric architecture as a Multi-Head structure, and introduce Multi-Head randomization to improve LLMs' performance on multi-task learning. The mathematical formalization of the Multi-Head structure is detailed in Section~\ref{multi-head}. Figure~\ref{architecture} illustrates the differences among the aforementioned structures.

However, in the Multi-Head architecture, the parameter similarity among head matrices remains high, hindering task-specific knowledge learning. This is due to the zero initialization of head matrices B, leading to similar update directions. To address this limitation, R-LoRA employs multi-head randomization, combining random initialization with multi-head dropout. This approach diversifies both the starting points and inputs of the head matrices, enabling more effective task-specific learning by breaking initial symmetry and promoting distinct optimization trajectories.
Our work makes the following key contributions: 

- \textit{We reveal redundancy and symmetry in the head matrices of Multi-Head LoRA, limiting its ability to capture diverse task-specific knowledge. } 
    
- \textit{We propose R-LoRA, introducing Multi-Head Randomization to enhance both performance and efficiency in multi-task learning.}
    
- \textit{Extensive experiments validate R-LoRA's superiority, with analysis showing performance gains stem from diversified head matrices.}

\section{Related Works}

\subsection{LoRA}
Current LLMs generally follow a decoder-only structure, characterized by a series of blocks, each comprising two key components with residual connections: a multi-head self-attention (MHA) layer and a feed-forward network (FFN)~\cite{vaswani2017attention}. These layers involve using dense learnable matrices. 

There is a need to adapt LLMs for specific tasks or domains with limited resources. To achieve this, low-rank adaptation (LoRA)~\cite{hu2021lora}, inspired by the concept of low intrinsic dimensionality in LLMs, decomposes the weight gradient $\Delta \mathbf{W}$ into low-rank matrices, thereby reducing the number of trainable parameters. Specifically, for a dense weight matrix $\mathbf{W} \in \mathbb{R}^{m \times n}$, LoRA employs two low-rank matrices, $\mathbf{B} \in \mathbb{R}^{m \times r}$ and $\mathbf{A} \in \mathbb{R}^{r \times n}$, to approximate the accumulated gradient updates $\Delta \mathbf{W}$. The rank $r$ is chosen to be much smaller than the minimum of $d$ and $k$, effectively decreasing the number of trainable parameters from $m \times n$ to $2r(m\times n)$. Consequently, the resulting weight matrix is expressed as $\mathbf{W} + \mathbf{B}\mathbf{A}$, and the output $h$ for an input $x$ through this updated weight matrix is formulated as:

\begin{equation}
    h = (\mathbf{W} + \Delta \mathbf{W}) x = \mathbf{W} x + \mathbf{B} \mathbf{A} x
    \label{eq:lora_output}
\end{equation}
Typically, matrix B is initialized with zeros, while matrix A is initialized using Kaiming Uniform~\cite{he2015delving}. This initialization strategy ensures that the initial outputs remain consistent with the pre-trained model, thereby avoiding the introduction of random disturbances.

Following LoRA, AdaLoRA~\cite{zhang2023adalora} dynamically learns the rank size needed for LoRA in each layer of the model. DeltaLoRA~\cite{zi2023delta} updates the original weights of the model using parameters from adapter layers, enhancing LoRA’s representational capacity. DoRA~\cite{liu2024dora} introduces a magnitude component to learn the scale of $\Delta W$ while utilizing the original AB as a direction component of $\Delta W$. PiSSA~\cite{meng2025pissa} and LoRA-GA~\cite{wang2024loragalowrankadaptationgradient} have improved the convergence speed and performance of LoRA by refining its initialization method. Their approaches focus on optimizing the initial parameter settings, which enhances the training dynamics and leads to more efficient and stable convergence. 

\subsection{Multi-Head architecture}
\label{multi-head}
MTL-LoRA~\cite{yang2024mtlloralowrankadaptationmultitask} and HydraLoRA~\cite{tian2024hydraloraasymmetricloraarchitecture} are pioneering methods that introduce the multi-head architecture into LoRA. This architecture is characterized by a central shared down-projection matrix A and multiple distinct head matrices B, enabling efficient and flexible adaptation across diverse tasks. 
As shown in Figure~\ref{architecture}, this architecture differentiates task-specific information while effectively capturing shared knowledge across various tasks. The Multi-Head architecture can be formulated as:

\begin{equation}
    W + \Delta W = W + \sum_{i=1}^{N} \omega_i \cdot B_i A
    \label{eq:multi-head}
\end{equation}

In MTL-LoRA~\cite{yang2024mtlloralowrankadaptationmultitask} and HydraLoRA~\cite{tian2024hydraloraasymmetricloraarchitecture}, the weights $w_i$ are computed through the routing matrix $W_r$ and the softmax function. It can be formulated as:
\begin{equation}
    \omega = Softmax(W_{r} x)
    \label{eq:multi-head}
\end{equation}

\subsection{Dropout}
Dropout is a widely used technique to prevent overfitting in deep networks by randomly deactivating units during training \cite{srivastava2014dropout}. This process samples from an exponential number of thinned networks, reducing unit co-adaptation and enhancing noise robustness. The following is the formulation of Dropout.
    
    
\begin{description}
  \item[\textbf{1. Mask vector:}]
    Generate a binary mask vector $\mathbf{m}\in\{0, 1\}^d$, where each element $m_j$ independently takes the value 1 with probability $1-p$ and 0 with probability $p$:  
    $ m_j \sim \text{Bernoulli}(p), \quad j=1,\ldots,d $

  \item[\textbf{2. Apply the mask:}]
    During training, multiply the input $\mathbf{x}$ or the activation values by the mask $\mathbf{m}$ element-wise to get the masked output $\tilde{\mathbf{x}}$:  
    $ \tilde{\mathbf{x}} = \mathbf{m} \odot \mathbf{x} $,  
    where $\odot$ denotes the Hadamard product (element-wise multiplication).

  \item[\textbf{3. Scale activation values:}]
    To maintain consistent expected outputs between training and testing, the retained neurons are typically scaled (multiplied by $\frac{1}{1-p}$):  
    $ \tilde{\mathbf{x}} = \frac{1}{1-p} \mathbf{m} \odot \mathbf{x} $
\end{description}
Dropout operations require both the computation and storage of masking vectors during training. At test time, the full network is utilized, benefiting from the ensemble effect of the thinned networks. In our work, we adapt dropout to a novel context within the multi-head structure of R-LoRA. Specifically, we employ dropout to differentiate the inputs of the head matrices, ensuring that each head learns distinct and complementary representations while also reducing computational overhead.

\begin{figure*}[t]
  \centering
  \includegraphics[width=0.8\linewidth]{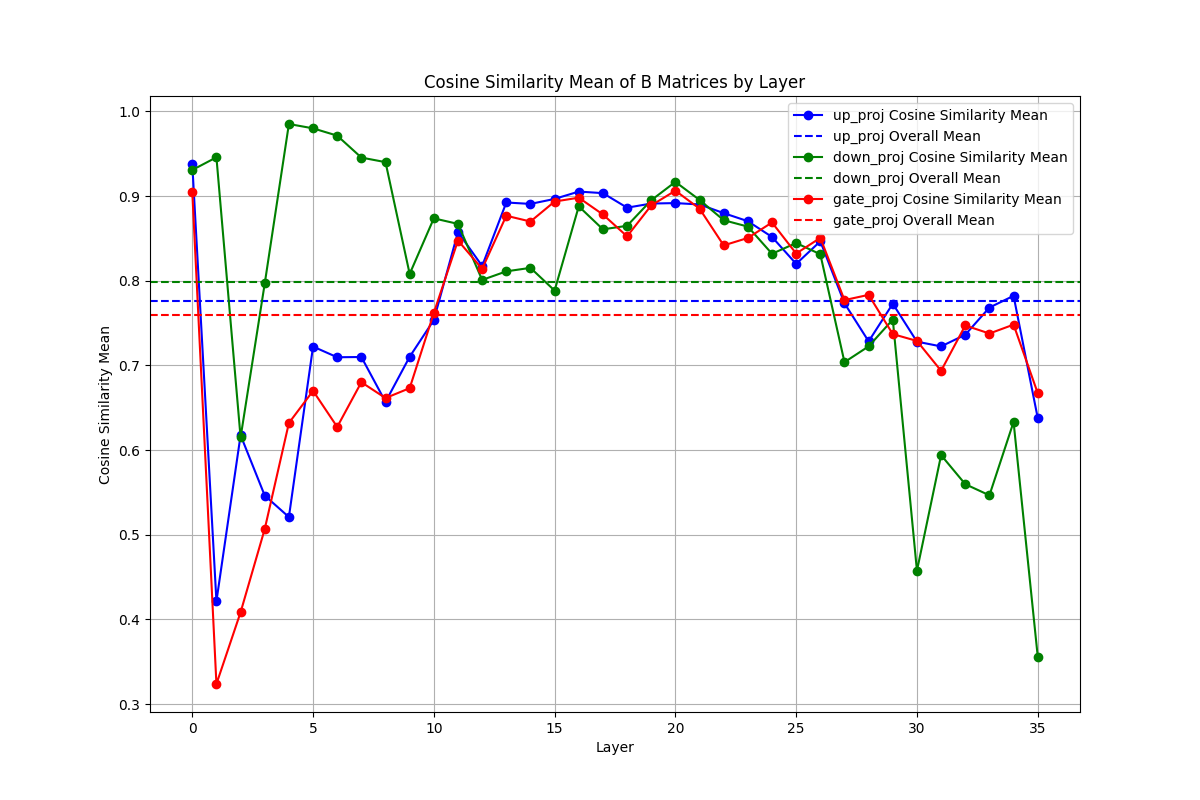}
  \caption {Cosine similarity among head matrices. "Overall mean" represents the average similarity across all layers.}
  \label{cosin}
\end{figure*}

\section{Motivation}
\label{Motivation}
In this section, we analyze the parameter similarity between different head matrices in the Multi-Head LoRA architecture. To achieve our objectives, we focus on HydraLoRA~\cite{tian2024hydraloraasymmetricloraarchitecture} and use cosine similarity and the T-SNE method to observe the parameters of the head matrices. We fine-tune Qwen2.5-3B \cite{qwen2.5} with HydraLoRA~\cite{tian2024hydraloraasymmetricloraarchitecture} on five different tasks. The details of the dataset can be referred to Appendix~\ref{motivation_data}. First, we flatten the head matrices into vectors and then calculate the cosine similarity between the vectors to obtain a similarity matrix. The average value of the matrix is regarded as the similarity of the head matrix corresponding to the parameter matrix. Additionally, we perform T-SNE analysis on all the head matrices in Figure~\ref{tsne} of Appendix~\ref{more result}. 

As shown in Figure~\ref{cosin}, the average similarity between different head matrices still reaches around 80\%. With such a high similarity, the knowledge learned between different head matrices is also quite similar, which hinders the learning of task-specific knowledge. To the best of our knowledge, this is due to the zero initialization of the head matrices. Tuning a pretrained LLM essentially becomes optimizing in a much smaller parameter space around the local optimum of pretrained models. After receiving the outputs from the shared down-projection matrix A, the outputs of the head matrices are highly similar in the early stages of training, leading to highly similar update directions during gradient updates.\\ \textbf{Research Question 1:} \textit{Is there a simple yet effective approach to differentiate head matrices such that they capture distinct task-specific knowledge for efficient multi-task learning?}

\begin{figure}[t]
  \includegraphics[width=\columnwidth]{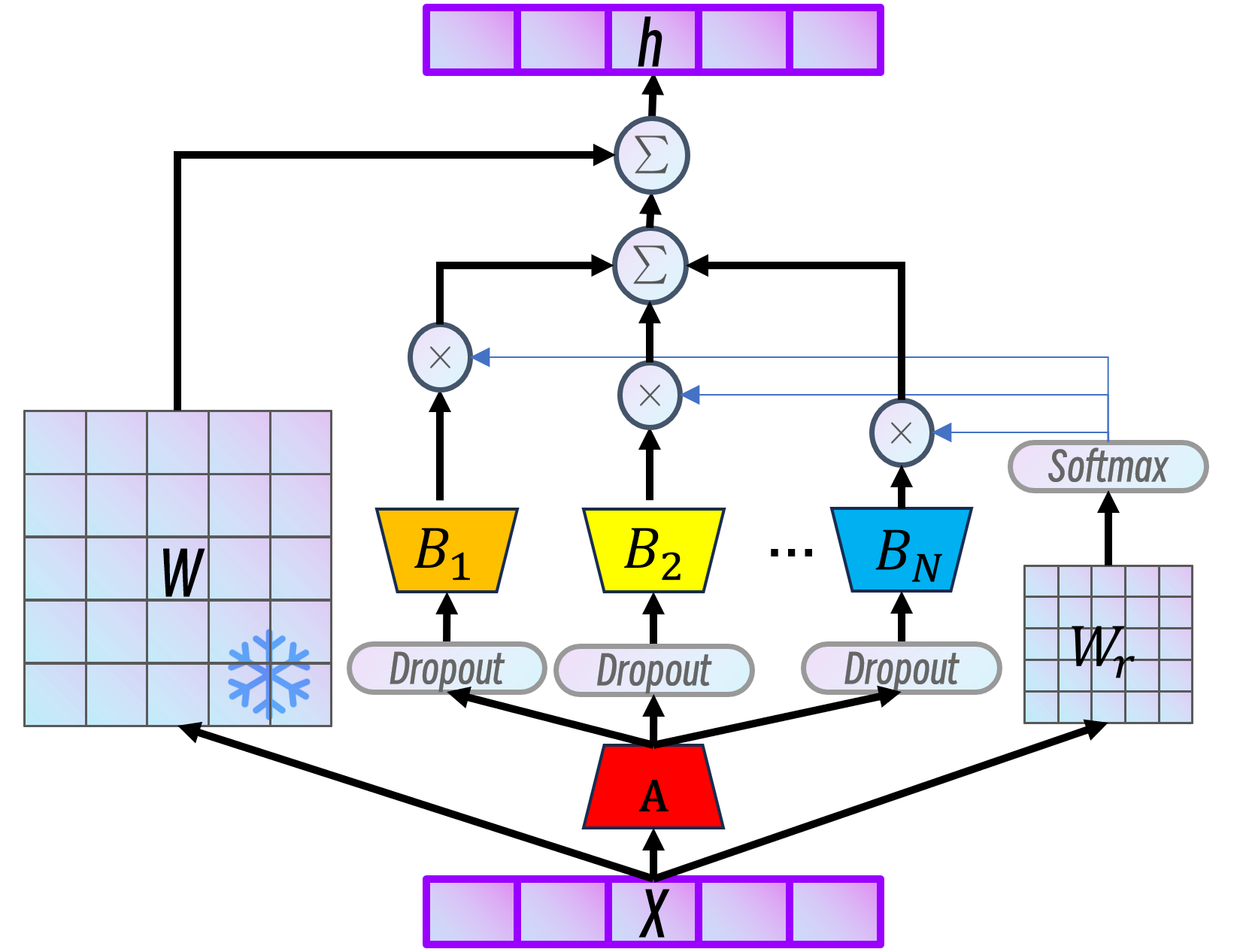}
  \caption{Overview of the R-LoRA.}
  \label{R-LoRA}
\end{figure}

\begin{table*}
  \centering
  \begin{tabular}{lcc}
    \hline
    \textbf{Method} & \textbf{A Init} & \textbf{B Init} \\
    \hline
    LoRA     & $U\left( -\sqrt{\frac{3}{d_{in}}}, \sqrt{\frac{3}{d_{in}}} \right)$ & $0$ \\
    HydraLoRA & $U\left( -\frac{1}{d_{in}}, \frac{1}{d_{in}} \right)$ & $0$ \\
    R-LoRA    & $\frac{\sqrt[4]{d_{out}}}{\sqrt{\gamma}} \cdot N\left( 0, \frac{1}{d_{in}} \right)$ & $\frac{\sqrt[4]{d_{out}}}{\sqrt{\gamma}} \cdot N\left( 0, \frac{1}{d_{out}} \right)$ \\
    \hline
  \end{tabular}
  \caption{Comparison of initialization.}
  \label{initialization}
\end{table*}

\section{Method}
In this work, we propose R-LoRA, which leverages multi-head randomization to assist the model in learning distinct knowledge. Multi-head randomization consists of two components: multi-head dropout and random initialization. An overview of R-LoRA is illustrated in Figure~\ref{R-LoRA}\\ \textbf{Research Objective: } \textit{To exploit randomization to differentiate the head matrices, thereby facilitating the optimization of their parameters to distinct regions and enhancing the diversity among the head matrices.}

\subsection{Multi-Head Dropout}
Multi-Head LoRA architecture is characterized by a shared down-projection matrix A and several distinct head matrices B. In HydraLoRA~\cite{tian2024hydraloraasymmetricloraarchitecture}, the head matrices receive the same output from the shared matrix A. According to ~\cite{hayou2024lora+} and~\cite{tian2024hydraloraasymmetricloraarchitecture}, the down-projection matrix A and the head matrix B in LoRA play distinct roles. Specifically, the down-projection matrix A primarily captures task-agnostic knowledge, encoding generalizable features applicable across tasks, while the head matrices specialize in task-specific knowledge, enabling the model to adapt to the unique requirements of individual tasks. This division of roles enhances the model's ability to balance generalization and specialization in multi-task learning scenarios. We propose employing multi-head dropout to differentiate the outputs of the down-projection matrix A, thereby ensuring that the head matrices produce distinct outputs. The framework of Multi-Head dropout and R-LoRA is shown in Figure~\ref{R-LoRA}. Our architecture builds upon the multi-head structure of LoRA by incorporating multi-head dropout. After the input is processed by the down-projection matrix A, it generates a task-agnostic representation. Multi-head dropout then diversifies this representation, allowing the model to learn task-specific knowledge from multiple perspectives and improving both generalization and task adaptability.

Additionally, R-LoRA's multi-head dropout mechanism offers practical advantages by reducing computational overhead and memory usage.  Unlike the original LoRA and Multi-Head structure LoRA, which perform dropout on the input $X \in \mathbb{R}^{b \times m}$, R-LoRA applies Multi-Head Dropout to the intermediate representations. $H \in \mathbb{R}^{b \times r},r \ll m$. Since dropout operations require both the computation and storage of masking matrices, applying dropout to the lower-dimensional $H$ results in reduced computational costs and lower GPU memory consumption. 

\subsection{Multi-Head Random Initialization}
The zero initialization of the head matrices results in identical starting points for the different head matrices during training, causing them to converge to similar positions. As demonstrated in Table~\ref{initialization}, To address this limitation, we adopt random initialization to break the symmetry of initial head matrices and diversify optimization trajectories, thereby encouraging them to converge to different positions. To stabilize the magnitude of outputs and enhance model performance, we incorporate a scaling coefficient into the initialization process of the head matrices. Specifically, inspired by \cite{he2015delving} and \cite{wang2024loragalowrankadaptationgradient}, we introduce a coefficient $\frac{\sqrt[4]{d_{out}}}{\sqrt{\gamma}}$ or $\frac{\sqrt[4]{d_{in}}}{\sqrt{\gamma}}$ during initialization to the matrices to ensure scale stability. The $\gamma$ is a hyperparameter set to 64 based on empirical findings from \cite{wang2024loragalowrankadaptationgradient}. Notably, \cite{wang2024loragalowrankadaptationgradient} theoretically analyzes that such a scaling factor helps maintain the numerical stability of the output magnitudes throughout training. When the head matrices are initialized with non-zero values, the initial $\Delta \mathbf{W_0}$ is no longer zero. To maintain consistency with the pre-trained model's initial outputs and avoid random perturbations, we subtract it from the original parameter matrix $\mathbf{W}$ during the initialization phase. It can be formulated as:

\begin{equation}
    W = W-\Delta \mathbf{W_0} = W-\frac{1}{N}\sum_{i=1}^{N} B_{i0} \cdot A_0
    \label{eq:multi-head}
\end{equation}

\begin{table*}[ht]
  \centering
  \begin{tabular}{l c c c c c c c c c c c c}
    \hline
    \textbf{Schemes} & \textbf{Task1} & \textbf{2} & \textbf{3} & \textbf{4}  & \textbf{5} & \textbf{6} & \textbf{7} & \textbf{8}& \textbf{Avg} & \textbf{\%Par} & \textbf{A} & \textbf{B} \\
    \midrule
    \multicolumn{6}{l}{\textit{Qwen2.5-3B}}\\
    \midrule
    LoRA*1  & 80.00 & 56.50 & 84.80 & 72.10 & 90.10 & 87.60 & 87.60 & 44.15 & 75.36 & 0.18 & 1 & 1 \\
    LoRA*2  & 86.30 & 56.40 & 84.70 & 72.60 & 91.40 & 87.90 & 87.60 & 44.80 & 76.46 & \textbf{0.45} & 1 & 1 \\
    Multi-LoRA  & 84.50 & 55.40 & 82.70 & 72.10 & 89.80 & 81.80 & 87.69 & 44.80 & 74.85 & 0.60 & 3 & 3 \\
    MoeLoRA  & 87.40 & 58.10 & 85.60 & 73.40 & 92.25 & 87.40 & 87.34 & 45.50 & 77.12 & 0.68 & 3 & 3 \\
    HydraLoRA & 86.50 & 56.40 & 85.00 & 73.40 & 92.00 & 87.40 & 88.38 & 45.10 & 76.77 & 0.45 & 1 & 3 \\
    R-LoRA  & 87.10 & 57.90 & 88.13 & 73.90 & 94.70 & 88.25 & 88.26 & 45.60 & \textbf{77.98} & \textbf{0.45} & 1 & 3 \\

    \midrule
    \multicolumn{6}{l}{\textit{Qwen2.5-7B}}\\
    \midrule

    LoRA*1 & 87.20 & 59.85 & 87.60 & 80.10 & 91.10 & 89.50 & 90.30 & 47.80 & 79.18 & 0.10 & 1 & 1 \\
    LoRA*2 & 88.40 & 60.80 & 88.40 & 81.50 & 93.60 & 91.20 & 91.80 & 48.10 & 80.48 & \textbf{0.25} & 1 & 1 \\
    Multi-LoRA & 88.30 & 58.90 & 87.50 & 79.80 & 91.50 & 88.40 & 91.90 & 47.90 & 79.28 & 0.33 & 3 & 3 \\
    MoeLoRA & 89.50 & 61.40 & 88.90 & 82.90 & 93.60 & 91.50 & 91.90 & 48.70 & 81.05 & 0.38 & 3 & 3 \\
    HydraLoRA & 88.60 & 61.20 & 89.50 & 81.70 & 93.60 & 91.60 & 91.70 & 48.10 & 80.75 & 0.25 & 1 & 3 \\
    R-LoRA & 89.80 & 62.50 & 89.40 & 83.70 & 95.10 & 92.10 & 92.17 & 50.80 & \textbf{81.95} & \textbf{0.25} & 1 & 3 \\

    \hline
  \end{tabular}
  \caption{Comparison of different training schemes on multi-task reasoning datasets. The rank of LoRA*2 was set to 10 to ensure that its trainable parameters matched those of R-LoRA, while all other configurations used a rank of 4.}
  \label{setting1}
\end{table*}

\begin{table*}
  \centering
  \begin{tabular}{l | cccccc}
    \hline
    \textbf{Metrics} & \textbf{Base} & \textbf{LoRA} & \textbf{LoRAHub*}& \textbf{LoRA MoE*}& \textbf{HydraLoRA}& \textbf{R-LoRA}\\
    \hline
    7B     & 31.6 & 37.1 & 39.7 & 40.3 & 41.5 & \textbf{42.2} \\
    13B & 38.4 & 40.8 & 41.9 & 43.7 & 44.2 & \textbf{45.1} \\
    A/B for training  & - & 1/1 & 48/48 & 48/48 & 1/10 & 1/10 \\
    A/B for inference & - & 1/1 & 20/20 & 48/48 & 1/10 & 1/10 \\
    \% Param & - & 0.062 & 1.240 & 2.976 & 0.341 & 0.341 \\
    \hline
  \end{tabular}
  \caption{Comparison of different training schemes on multi-tasks. * indicates results from \cite{tian2024hydraloraasymmetricloraarchitecture}.}
  \label{multi}
\end{table*}

\begin{table*}
  \centering
  \begin{tabular}{l | cccc}
    \hline
    \textbf{Schemes} & \textbf{3 Heads(bfloat16)} & \textbf{5 Heads(bfloat16)} & \textbf{3 Heads(float32)} & \textbf{5 Heads(float32)}\\
    \hline
    MD & 18.53GB / 2.20h & 22.05GB / 2.75h & 34.23GB / 8.68h & 41.24GB / 9.65h \\
    ID       & 23.42GB / 2.41h & 30.25GB / 3.25h & 42.09GB / 9.08h & 54.45GB / 10.31h  \\
    \hline
  \end{tabular}
  \caption{Comparison of memory consumption and per-epoch training time across different dropout operations. MD denotes our proposed Multi-Head Dropout, while ID represents input dropout applied to x in HydraLoRA.}
  \label{gpu}
\end{table*}

\begin{figure*}[t]
  \centering
  \includegraphics[width=0.8\linewidth]{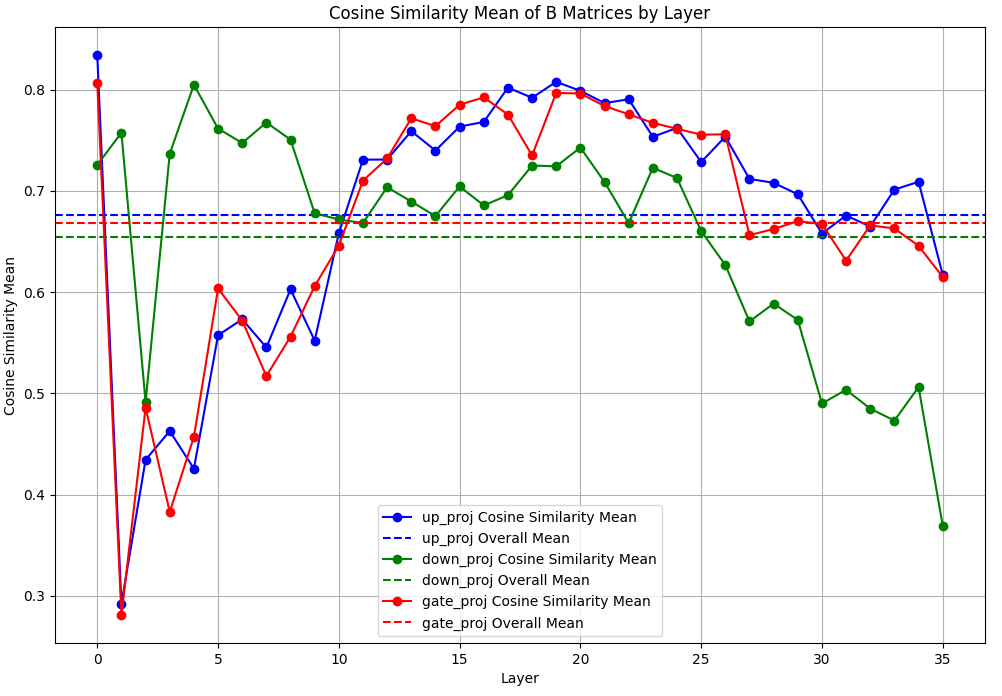}
  \caption {Cosine similarity among head matrices in R-LoRA. "Overall mean" represents the average similarity across all layers.}
  \label{cosin_mt}
\end{figure*}

\begin{figure}[t]
  \includegraphics[width=\columnwidth]{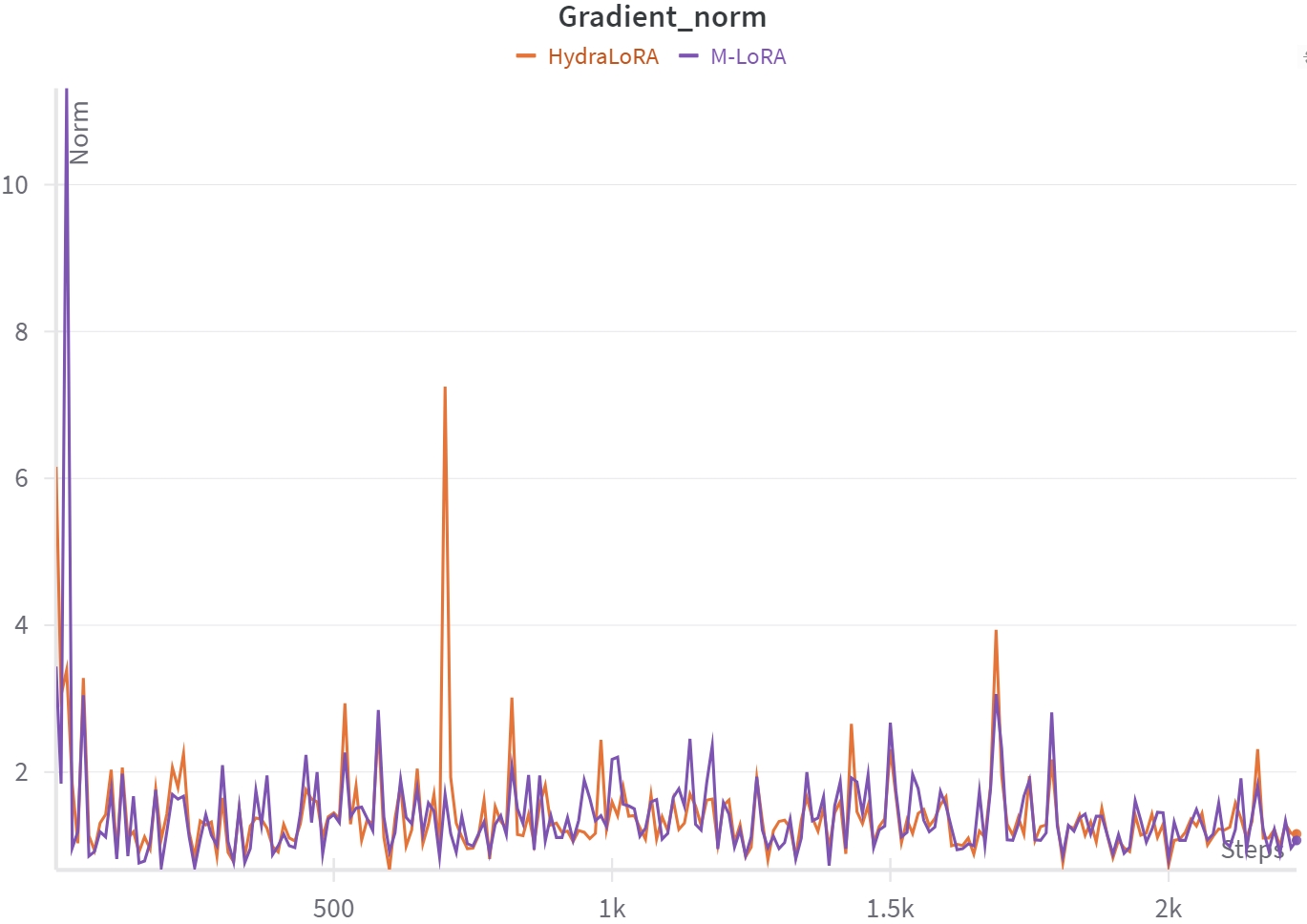}
  \caption{Gradient norm dynamics during training: Comparison of conventional and multi-head randomized configurations, highlighting enhanced stability through diversified head matrices.}
  \label{Gradient}
\end{figure}

\section{Experiments}

In this section, we validate the effectiveness of R-LoRA across various models and settings. Specifically, we evaluate R-LoRA's multi-task adaptability using Qwen2.5~\cite{qwen2.5} in Setting 1 and its multi-task generalization capability using LLaMA-2~\cite{touvron2023llama} in Setting 2. Model sizes range from 3B to 13B. An extensive ablation study further demonstrates the effectiveness of multi-head randomization in R-LoRA.

\subsection{Experiment Settings}

\noindent\textbf{Datasets \& Benchmarks:}  
\textbf{Setting 1:} We fine-tune Qwen2.5 on datasets covering commonsense and mathematical reasoning tasks and evaluate performance on their respective test sets.  
\textbf{Setting 2:} We fine-tune LLaMA-2 on a subset of the Flanv2 dataset~\cite{brown2020language}, which includes tasks grouped into 10 distinct task clusters. Performance is measured using the Big-Bench Hard (BBH) benchmark. Additional details about the datasets and implementation details can be found in Appendix \ref{appendix} and Appendix~\ref{implementation}, respectively.

\noindent\textbf{Baselines:}  
In Setting 1, we compare R-LoRA with LoRA, Multi-LoRA, MoeLoRA, and HydraLoRA.  
In Setting 2, the comparison includes LoRAHub~\cite{huang2023lorahub}, which employs black-box optimization for weighted averaging of LoRAs, LoRA-MoE~\cite{liu2024moemeetsllmsparameter}, which integrates lightweight experts with a Mixture of Experts architecture, and HydraLoRA.

\subsection{Performance of R-LoRA on Multi-Tasks}
The evaluation across diverse multi-task reasoning datasets, as shown in Table~\ref{setting1} and Table~\ref{multi}, demonstrates that R-LoRA achieves superior performance compared to all other methods. By introducing multi-head randomization, R-LoRA achieves significantly improved multi-task adaptability and generalization capabilities. The performance gains achieved by R-LoRA, driven by these innovations, outperform LoRA and SOTA multi-head LoRA methods like HydraLoRA. This highlights R-LoRA's enhanced generalization and task adaptability. Additional results on the performance of R-LoRA under single-task settings are provided in Appendix~\ref{more result}.

\subsection{Efficiency of R-LoRA}
We conducted a comparative analysis of the memory usage and training time between R-LoRA and the original Multi-head structure LoRA using Qwen2.5-3B under various configurations. Table~\ref{gpu} demonstrates that R-LoRA's multi-head dropout approach reduces GPU memory consumption by up to 20\% and cuts training time by up to 8\%, highlighting its superior efficiency in comparison to traditional methods.

\begin{table*}[ht]
  \centering
  \begin{tabular}{l | ccccccccc}
    \toprule
    \textbf{Schemes} & \textbf{Task1} & \textbf{2} & \textbf{3} & \textbf{4} & \textbf{5} & \textbf{6} & \textbf{7} & \textbf{8} & \textbf{Avg}\\
    \midrule
    \multicolumn{6}{l}{\textit{Llama3.2-3B}}\\
    \midrule
    R-LoRA     & 95.82 & 83.68 & 84.25 & 85.48 & 71.45 & 74.12 & 84.34 & 85.39 & \textbf{83.07} \\
    w/o MD   & 95.24 & 83.25 & 82.46 & 84.75 & 70.27 & 73.96 & 84.57 & 83.29 & 82.22 \\
    w/o MI   & 94.66 & 82.77 & 83.58 & 83.25 & 69.79 & 74.23 & 83.28 & 84.16 & 81.97 \\
    Zero A     & 95.67 & 83.46 & 83.47 & 85.64 & 70.27 & 73.84 & 84.13 & 84.89 & \underline{82.67} \\
    HydraLoRA     & 95.12 & 82.14 & 83.88 & 82.68 & 69.86 & 72.33 & 79.25 & 84.25 & 81.19 \\
    \midrule
    \multicolumn{6}{l}{\textit{Qwen2.5-3B}}\\
    \midrule
    R-LoRA   & 96.42 & 83.27 & 85.34 & 86.49 & 72.84 & 73.86 & 86.24 & 88.94 & \textbf{84.18} \\
    w/o MD   & 96.22 & 83.66 & 83.25 & 84.72 & 71.15 & 73.24 & 85.02 & 88.12 & 83.17 \\
    w/o MI   & 96.10  & 83.21 & 83.65 & 84.50  & 71.69 & 72.05 & 83.24 & 88.36 & 82.85 \\
    Zero A      & 96.24 & 84.02 & 84.36 & 85.89 & 72.13 & 73.51 & 85.26 & 89.13 & \underline{83.82} \\
    HydraLoRA       & 95.89 & 83.53 & 82.97 & 84.24 & 70.95 & 71.93 & 83.06 & 87.33 & 82.49 \\
    \bottomrule
  \end{tabular}
  \caption{Results of Ablation Studies on Qwen2.5 and Llama3.2 with Different Schemes Across Various Tasks. The table compares R-LoRA with its ablated versions (without Multi-Head Dropout/MD, without Multi-Head Initialization/MI, and zero initialization to LoRA A in R-LoRA) against HydraLoRA across eight tasks.}
  \label{8-task}
\end{table*}

\subsection{Parameter Analysis}
\textbf{Research Question2:} \textit{Does multi-head randomization effectively enhance the acquisition of diverse knowledge across the head matrices?}

In this section, we analyze the parameter differences among the head matrices in R-LoRA. The methodology and experimental setup align with those described in Section~\ref{Motivation}. As shown in Figure~\ref{cosin_mt}, the parameter similarity between head matrices in R-LoRA is reduced to below 70\%. This significant decrease indicates that multi-head randomization effectively enhances the model's capacity to learn task-specific knowledge, thereby mitigating redundant learning and increasing the diversity of acquired knowledge across tasks.

\subsection{Training Process}
\textbf{Research Question3:} \textit{Does multi-head randomization impact the stability of the training process?}

As illustrated in Figure~\ref{Gradient}, R-LoRA benefits from multi-head randomization, exhibiting significantly larger gradient norms in the early stages of training compared to HydraLoRA. This drives the head matrices to converge to distinct regions, enhancing the model's ability to capture diverse representations and improving overall performance. Furthermore, R-LoRA exhibits superior training stability, as evidenced by its more stable gradient norms throughout the training process. This stability enables the model to effectively acquire diverse knowledge without compromising training efficiency and robustness.

\subsection{Ablation Study}
Ablation studies were conducted on Llama3.2-3B and Qwen2.5-3B models across eight-task configurations, using 11 datasets spanning 8 categories. All models were evaluated on their respective test sets, with results summarized in Table~\ref{8-task}. Dataset details are provided in the Appendix \ref{ablation}. More results on the smaller model Qwen2.5-0.5B are shown in the Appendix~\ref{more result}

Experimental results demonstrate that the two key components of multi-head randomization—random initialization and dropout—are pivotal for enhancing the model's adaptability across tasks. Multi-head randomization remains effective even when initialization is isolated to LoRA B. As shown in Table~\ref{8-task}, R-LoRA with zero-initialized LoRA A (Zero A) consistently outperforms HydraLoRA, demonstrating that the performance gains are primarily attributed to the diversified parameter spaces in the head matrices B. Random initialization assigns unique weights to each head matrix, enabling task-specific pattern capture. Dropout diversifies inputs to the head matrices, promoting distinct learning pathways. Together, these components improve task-specific feature capture while maintaining robustness in multi-task learning.

\section{Discussion}
In this section, we discuss the underlying mechanisms behind R-LoRA’s multi-head randomization that enhance multi-task learning. As revealed in the Motivation section, a key limitation of traditional multi-head LoRA lies in the high similarity among head matrices, which stems from zero-initialization. This initialization scheme confines heads to symmetric states, limiting their ability to explore sparse yet critical subspaces, such as those relevant to rare syntactic relationships. Consequently, heads tend to converge on overlapping subspaces, failing to adequately address task-specific requirements.

Each token’s semantic logics (e.g., semantic, syntactic, contextual) naturally reside in multiple subspaces of high-dimensional embedding spaces. The Multi-Head mechanism excels by decomposing these logics into distinct subspaces for independent processing. R-LoRA aligns with this principle through two key innovations:  

- \textbf{Multi-Head Dropout}: Promotes heterogeneous feature learning to capture complementary aspects of the embedding space.

- \textbf{Multi-Head Random Initialization}: Decouples head trajectories to prevent convergence to overlapping subspaces.

As shown in Figure~\ref{cosin_mt}, R-LoRA effectively reduces the similarity among head matrices, promoting diverse feature learning across tasks. Empirical results in Table \ref{8-task} further confirm that the performance gains are primarily attributed to the diversified parameter spaces in LoRA B, rather than LoRA A, highlighting the importance of the up-projection module in enabling task-specific adaptation.

\section{Conclusion}
In this work, we first analyze the multi-head structure of LoRA, revealing excessive similarity among head matrices that limits task-specific learning. To address this, R-LoRA introduces multi-head randomization, a simple yet effective approach that differentiates head matrices, enabling the model to learn diverse knowledge across tasks. This innovation enhances both performance and efficiency, reducing GPU memory usage and training time.

Extensive experiments validate R-LoRA's superiority. The performance gains stem primarily from increased diversity in the parameter spaces of the head matrices, confirming the effectiveness of R-LoRA for efficient multi-task learning.
\section{Limitation}
Despite the promising results of R-LoRA, several limitations should be acknowledged. While we have conducted extensive experiments to validate its effectiveness, the inherent complexity of multi-task learning highlights the importance of further exploration and broader evaluation. Currently, our validation focuses on NLP tasks, and extending the method to other modalities, such as computer vision and multimodal settings, represents an exciting avenue for future research. These directions could help unlock the full potential of R-LoRA and deepen our understanding of its applicability across diverse domains.



\bibliography{custom}

\appendix


\section{More Results}
\label{more result}

\subsection{T-SNE analysis}
\begin{figure*}[t]
  \centering
  \includegraphics[width=0.32\linewidth]{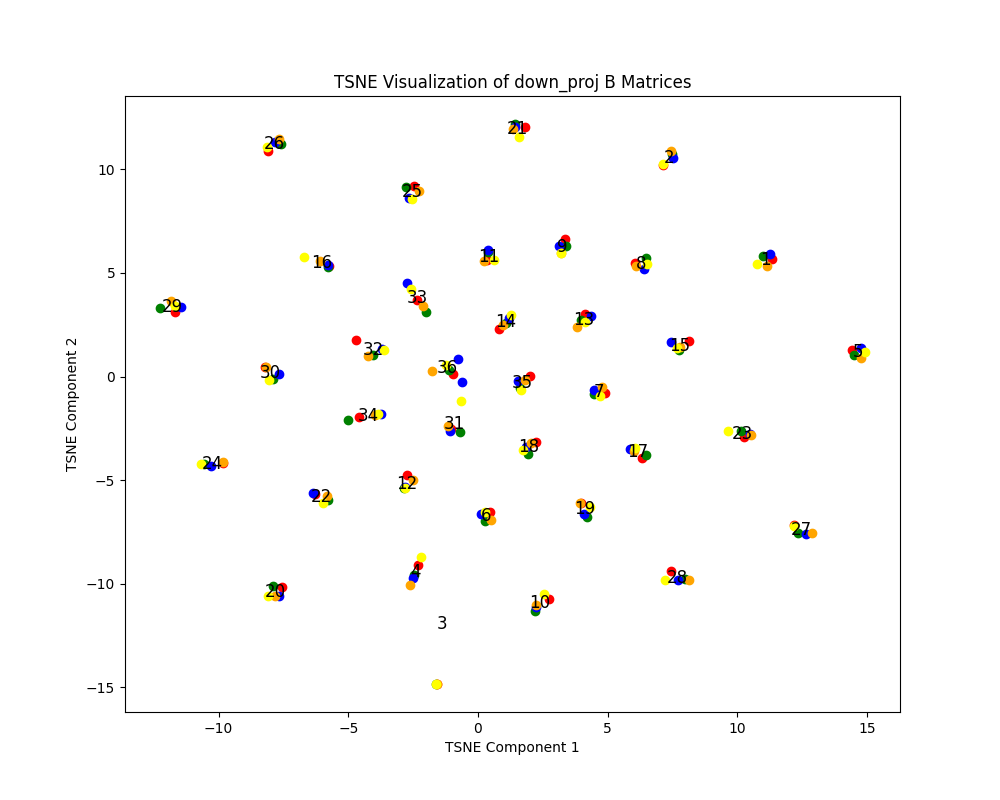}
  \includegraphics[width=0.32\linewidth]{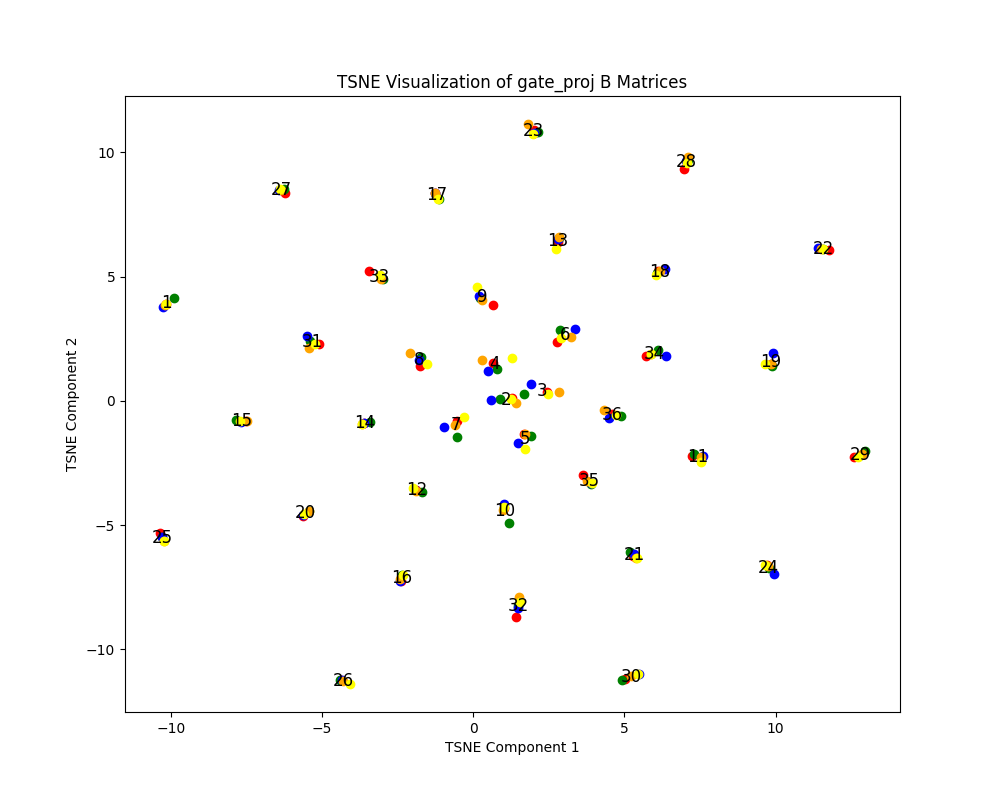}
  \includegraphics[width=0.32\linewidth]{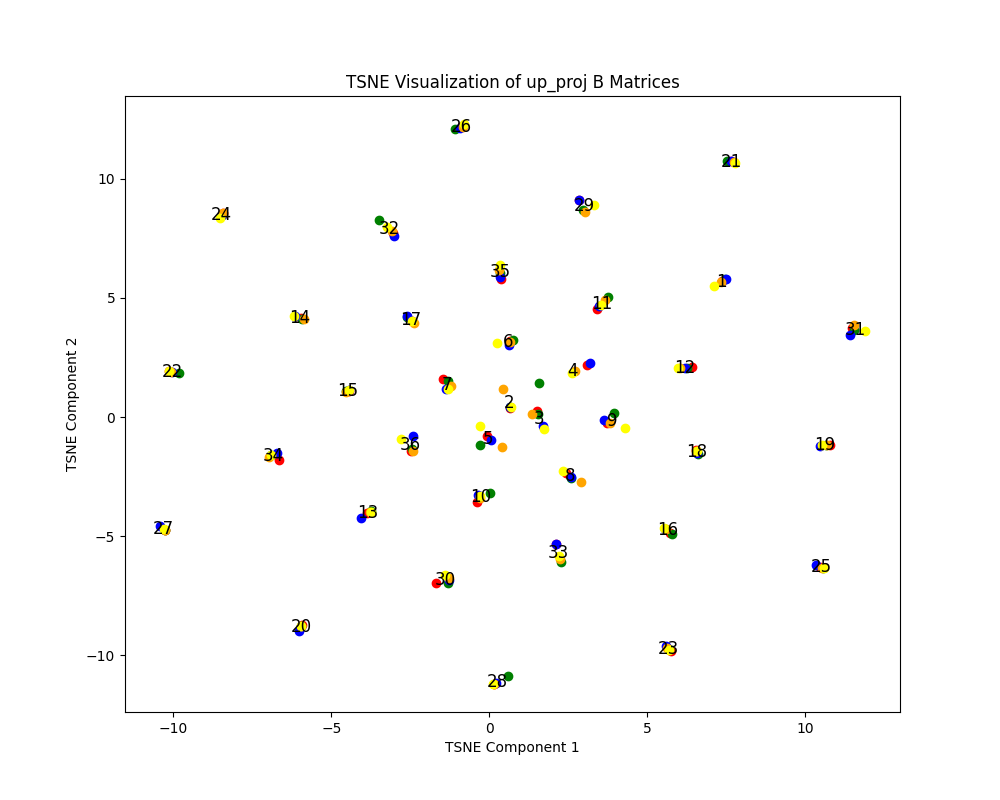}
  \caption {T-SNE analysis of head matrices in HydraLoRA}
  \label{tsne}
\end{figure*}

 The T-SNE analysis of head matrices in HydraLoRA is shown in Figure~\ref{tsne}.

\subsection{Performance of R-LoRA on Single Task}
We compare R-LoRA against various PEFT methods on single datasets: 1) Full fine-tuning; 2) Prompt Tuning~\cite{lester2021power}; 3) P-Tuning~\cite{liu2024gpt}; 4) Prefix Tuning~\cite{li2021prefix}; 5)  \(IA^{3}\)~\cite{liu2022few}; 6) AdaLoRA~\cite{zhang2023adalora}; 7) HydraLoRA~\cite{tian2024hydraloraasymmetricloraarchitecture}. 

As shown in Table~\ref{single}, in the single-task setting, where the knowledge and text format of the data are relatively homogeneous, R-LoRA demonstrates slightly improved performance compared to HydraLoRA. While multi-head randomization is primarily designed for multi-task learning, its ability to learn diverse knowledge remains beneficial even in single-task scenarios. This slight edge over HydraLoRA underscores R-LoRA's capacity to capture varied patterns effectively, even when its full potential is not fully utilized in single-dataset settings. These results further highlight R-LoRA's robustness and adaptability across different task complexities.

\subsection{Ablation study of R-LoRA on smaller model}
\label{Qwen2.5-0.5B}
Table~\ref{5-task} show the ablation study on Qwen2.5-0.5B

\subsection{Datasets in Single-task}
\begin{enumerate}
    \item \textbf{General}: We fine-tune with the general instruction tuning dataset \texttt{databricks-dolly-15k} for generic language capability and evaluate with MMLU.
    
    \item \textbf{Medical}: We fine-tune with \texttt{GenMedGPT} and \texttt{clinic-10k} from \texttt{ChatDoctor} for medicine applications and evaluate medical tasks in MMLU including three related tasks: "clinical knowledge", "professional medicine", and "college medicine".
    
    \item \textbf{Law}: We fine-tune with two legal instruction tuning datasets \texttt{Lawyer-Instruct} and \texttt{US-Terms} then evaluate with law tasks in MMLU including two related tasks: "professional law" and "international law".
    
    \item \textbf{Math}: We fine-tune with the training split of \texttt{GSM8K} for mathematical reasoning and evaluate with the test set of \texttt{GSM8K}.
    
    \item \textbf{Code}: We fine-tune with \texttt{CodeAlpaca} for code generation and evaluate with \texttt{HumanEval}.
\end{enumerate}

\begin{table*}[ht]
  \centering
  \begin{tabular}{l c c c c c c c c c c c}
    \hline
    \textbf{Schemes} & \textbf{General} & \textbf{Medical} & \textbf{Law} & \textbf{Code}  & \textbf{Math}  & \textbf{Avg} & \textbf{\%Param} & \textbf{\#A} & \textbf{\#B} \\
    \hline
    Base*  & 38.88 & 35.98 & 33.51 & 20.34 & 10.38 & 27.82 & - & - & -  \\
    Full* & 49.91 & 46.78 & 46.08  & 32.93 & 25.70 & 40.28 & 100 & - & -  \\
    \hline
    Prompt Tuning*  & 39.91 & 37.59 & 35.02  & 21.55 & 13.18 & 29.45 & 0.001 & - & -  \\
    P-Tuning*  & 41.11 & 39.81 & 36.72  & 21.13 & 15.56 & 30.87 & 0.193 & - & -  \\
    Prefix Tuning*  & 41.78 & 40.28 & 36.54  & 22.56 & 16.89 & 31.61 & 0.077 & - & -  \\
     \(IA^{3}\)*  & 40.45 & 37.12 & 35.25  & 23.17 & 13.98 & 29.99 & 0.009 & - & - \\
    \hline
    LoRA$(r=8)$  & 43.44 & 41.18 & 37.95 & 22.82 & 18.72 & 32.82 & 0.062 & 1 & 1 \\
    AdaLoRA*$(r=8)$  & 44.32 & 42.83 & 39.36 & 23.78 & 19.51 & 33.96 & 0.093 & 1 & 1 \\
    LoRA$(r=16)$  & 45.12 & 43.22 & 40.24 & 25.22 & 20.14 & 34.79 & 0.124 & 1 & 1 \\
    HydraLoRA$(r=8)$  & 46.89 & 45.21 & 42.88 & 27.43 & 22.27 & 36.94 & 0.124 & 1 & 3 \\
    \hline
    R-LoRA$(r=8)$  & 47.02 & 45.54 & 43.23 & 27.27 & 22.12 & \textbf{37.04} & 0.124 & 1 & 3 \\
    \hline
  \end{tabular}
  \caption{Comparison of different training schemes on single task. * indicates results from \cite{tian2024hydraloraasymmetricloraarchitecture}}
  \label{single}
\end{table*}

\begin{table*}
  \centering
  \begin{tabular}{l | cccccc}
    \hline
    \textbf{Schemes} & \textbf{Task1} & \textbf{2} & \textbf{3} & \textbf{4} & \textbf{5} & \textbf{Avg}\\
    \hline
    R-LoRA          & \textbf{91.74} & \textbf{81.50} & \textbf{77.60} & \textbf{67.80} & 49.30 & \textbf{73.59} \\
    w/o MD   & 91.40 & 81.20 & 77.10 & 66.10 & 49.10 & 72.98  \\
    w/o MI   & 91.20 & 80.80 & 77.50 & 66.20 & \textbf{49.40} & 73.02  \\
    HydraLoRA  & 90.97 & 80.30 & 77.20 & 65.80 & 49.20 & 72.69 \\
    \hline
  \end{tabular}
  \caption{Results of Ablation Studies on Qwen2.5-0.5B with Different Schemes Across Various Tasks. The table compares R-LoRA with its ablated versions (without Multi-Head Dropout/MD and without Multi-Head Random Initialization/MI) against HydraLoRA across five tasks.}
  \label{5-task}
\end{table*}

\section{Datasets}
\label{appendix}
\subsection{Motivation}
\label{motivation_data}

In the section of Motivation, We fine-tune Qwen2.5-3B on five tasks: Paraphrase Detection (QQP), Natural Language Inference (QNLI)~\cite{wang2018glue}, Commonsense Reasoning (SIQA)~\cite{sap2019socialiqa}, Physical Commonsense Reasoning (PIQA)~\cite{bisk2020piqa}, and Math (GSM8K)~\cite{cobbe2021training}

\subsection{Setting 1}
\begin{enumerate}
    \item \textbf{Reading Comprehension}: BoolQ
    \item \textbf{Science Question Answering}: SiQA
    \item \textbf{Physical Question Answering}: PiQA
    \item \textbf{Word Relation Reasoning}: Winogrande
    \item \textbf{Commonsense Reasoning}: Hellaswag
    \item \textbf{Open-Book Question Answering}: OBQA
    \item \textbf{Closed-Book Question Answering}: ARC
    \item \textbf{Mathematical Reasoning}: GSM8K
\end{enumerate}

\subsection{Setting 2}
Following~\cite{tian2024hydraloraasymmetricloraarchitecture}, for complex mixed multi-task/domain, we select a portion of the \texttt{Flanv2} datasets covering Natural Language Understanding (NLU) and Natural Language Generation (NLG), which can be grouped into 10 distinct task clusters. Then we evaluate it with the Big-Bench Hard (BBH) benchmark.

We summarize the details of the used datasets as follows:

\begin{enumerate}
    \item \textbf{Struct-to-Text Conversion}: This task evaluates the capability to generate natural language descriptions from structured data inputs. We use the following datasets: (1) CommonGen; (2) DART; (3) E2ENLG; (4) WebNLG
    
    \item \textbf{Translation}: Translation involves converting text from one language to another, maintaining the original meaning and nuances. We use the following datasets: (1) En-Fr from WMT'14; (2) En-De, En-Tr, En-Ru, En-Fi, En-Ro from WMT'16; (3) En-Es from Paracrawl.
    
    \item \textbf{Commonsense Reasoning}: This involves assessing the ability to apply physical or scientific principles alongside common sense in reasoning tasks. We use the following datasets: (1) COPA; (2) HellaSwag; (3) PiQA; (4) StoryCloze.
    
    \item \textbf{Sentiment Analysis}: A fundamental task in natural language processing (NLP) that determines the sentiment polarity (positive or negative) of a given text. We use the following datasets: (1) IMDB; (2) Sentiment140; (3) SST-2; (4) Yelp.
    
    \item \textbf{Paraphrase Detection}: This task requires models to ascertain whether two sentences convey the same meaning, indicating semantic equivalence. We use the following datasets: (1) MRPC; (2) QQP; (3) Paws Wiki.
    
    \item \textbf{Coreference Resolution}: Involves identifying instances within a text that refer to the same entity, demonstrating an understanding of textual context. We use the following datasets: (1) DPR; (2) WSC273.
    
    \item \textbf{Reading Comprehension}: Assesses the capability to derive answers to questions from a provided text containing relevant information. We use the following datasets: (1) BoolQ; (2) DROP; (3) MultiRC; (4) OBQA; (5) SQuADv1; (6) SQuADv2.
    
    \item \textbf{Reading Comprehension with Commonsense}: Merges traditional reading comprehension skills with commonsense reasoning, requiring understanding beyond the explicit text. We use the following datasets: (1) CosmosQA; (2) ReCoRD.

    \item \textbf{Natural Language Inference}: Focuses on deducing the relationship between two sentences, determining if the second sentence logically follows from, contradicts, or is unrelated to the first sentence. We use the following datasets: (1) ANLI; (2) CB; (3) MNLI; (4) QNLI; (5) SNLI; (6) WNLI; (7) RTE.
    
    \item \textbf{Closed-Book Question Answering}: This task challenges models to answer questions about general knowledge without direct access to external information sources. We use the following datasets: (1) ARC; (2) NQ; (3) TriviaQA.

\end{enumerate}

\subsection{Ablation Study}
\label{ablation}
Due to limited computational resources, we selected a subset of the dataset for training and testing.
\textbf{Five tasks for Smaller model Qwen2.5-0.5B in Appendix~\ref{Qwen2.5-0.5B}}:
    \begin{itemize}
        \item Task 1: Sentiment Analysis (SST2)
        \item Task 2: Paraphrase Detection (QQP)
        \item Task 3: Natural Language Inference (QNLI)
        \item Task 4: Physical Commonsense Reasoning (PiQA)
        \item Task 5: Commonsense Reasoning (SiQA)
    \end{itemize}

\textbf{Eight tasks}:
    \begin{itemize}
        \item Task 1: Sentiment Analysis (SST2)
        \item Task 2: Paraphrase Detection (QQP)
        \item Task 3: Natural Language Inference (MNLI + QNLI)
        \item Task 4: Reading Comprehension (BoolQ + OBQA)
        \item Task 5: Commonsense Reasoning (PiQA + SiQA)
        \item Task 6: Reading Comprehension with Commonsense (CosmosQA)
        \item Task 7: Coreference Resolution (Winogrande)
        \item Task 8: Closed-Book Question Answering (ARC)
    \end{itemize}

\section{Implementation Details}
\label{implementation}
The hyperparameters used for training are as follows: a learning rate of 0.0002, "lora\_alpha"=32, and trainable LoRA components including "gate\_proj", "down\_proj", and "up\_proj". A dropout rate of 0.2 was applied to the LoRA, with a warmup ratio of 0.03. Mixed-precision training was enabled using bfloat16, and the learning rate scheduler was set to cosine annealing. The model was trained for 1 epoch on NVIDIA 4090 GPU. In Setting 1, the rank of LoRA was set to 10 for LoRA*2 to match the total number of trainable parameters in R-LoRA, while the rank for others was set to 4.


\begin{figure}[t]
  \includegraphics[width=\columnwidth]{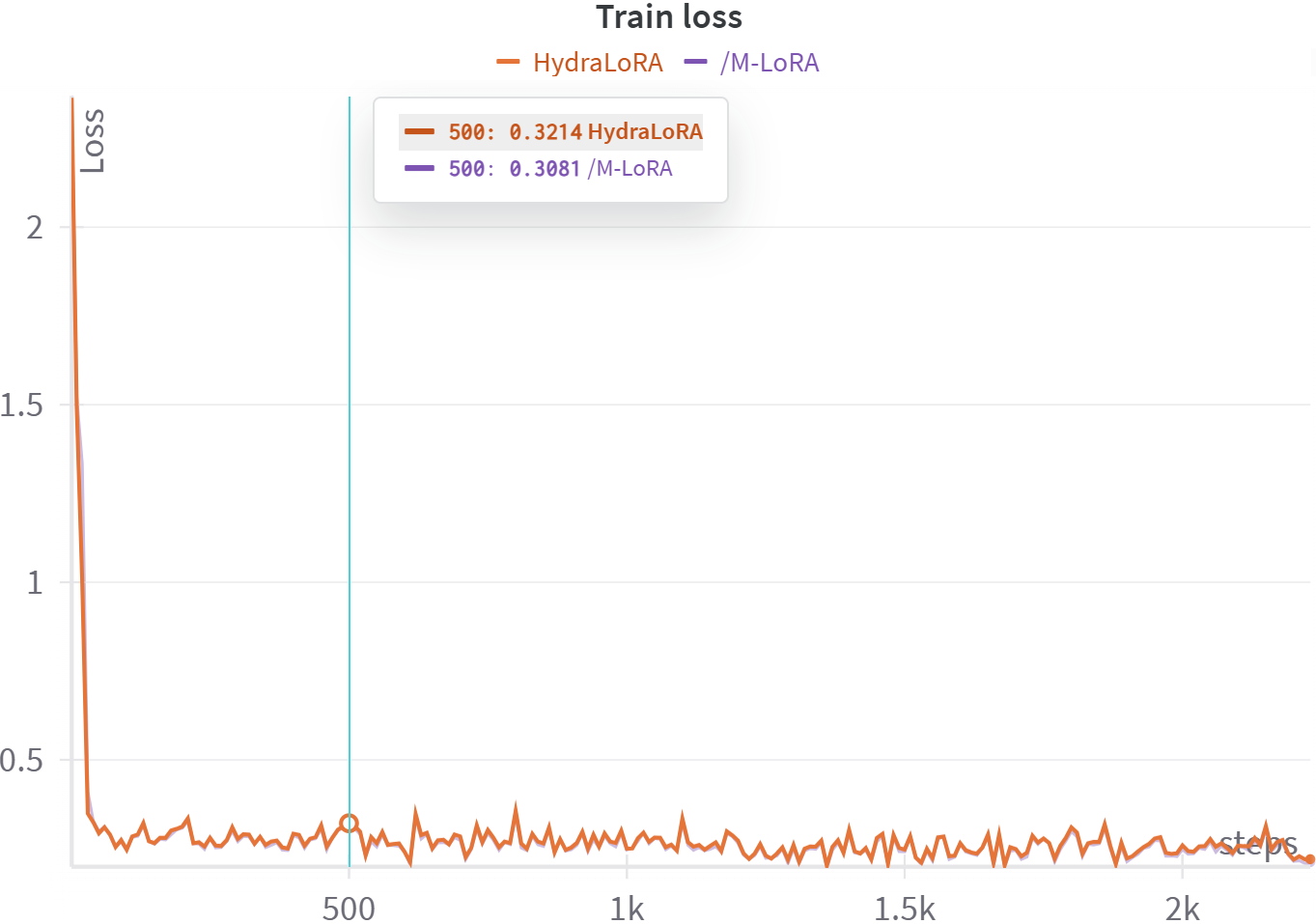}
  \caption{Training loss curves of HydraLoRA and R-LoRA. The loss of R-LoRA remains lower throughout the entire training process.}
  \label{loss}
\end{figure}

\section{Baselines}
\begin{enumerate}

    \item \textbf{Prompt Tuning}: This method adds task-specific prompts to the input. These prompt parameters are updated independently while the pretrained model parameters remain frozen.

    \item \textbf{P-Tuning}: This method incorporates trainable prompt embeddings into the input, optimized by a prompt encoder to automatically discover effective prompts, removing the need for manual design. Prompt tokens can be placed anywhere in the input sequence, and anchor tokens are introduced to enhance performance.

    \item \textbf{Prefix Tuning}: This method prefixes a series of task-specific vectors to the input sequence. These prefix parameters can be learned while keeping the pretrained model frozen. The prefix parameters are inserted into all layers of the model.

    \item \textbf{\(IA^{3}\)}: This method enhances efficiency by infusing learned vectors into transformer architectures, drastically reducing the number of trainable parameters.

    \item \textbf{AdaLoRA}: Unlike LoRA, which distributes parameters evenly across all modules, AdaLoRA optimizes the number of trainable parameters assigned to weight matrices and layers. More parameters are allocated to important weight matrices and layers, while less important ones receive fewer parameters.

    \item \textbf{LoraHub} randomly aggregates 20 LoRAs for new downstream tasks. It employs a black-box optimization technique to determine the weight of each LoRA, eliminating the need for gradient calculations of the large model. This involves parameter-level weighted averaging.
    
    \item \textbf{LoRA MoE}. A collection of $n$ parameterized experts, denoted as $E_1, \ldots, E_n$, is orchestrated by a router network $R$. $E_i=B_iA_i$. Router network features a dense layer with adjustable weights $W_R$ from $\mathbb{R}^{d_m \times n}$. A softmax function then processes an intermediate token representation $x$, yielding gating scores $s_1, \ldots, s_n$ that determine the weighted contribution of each expert's output:
\begin{equation}
    s_i = R(x)_i = \text{softmax}(Top(W_R^T x, K))
\end{equation}
Subsequently, the overall output $y$ is synthesized by aggregating the Top-K experts' outputs, each modulated by its respective gating score:
\begin{equation}
    y = \sum_{i=1}^n s_i \cdot E_i(x) \quad (\text{MoE})
\end{equation}
This results in a dynamic allocation of the model's capacity, enabling specialized processing by experts as directed by the router's gating mechanism.

    \item \textbf{HydraLoRA} uses a shared matrix $A$ and multiple matrices $B_1, \ldots, B_n$. The shared matrix $A$ is used to project the input vector $x$ into a lower-dimensional space, while each matrix $B_i$ is used to modulate the output of the corresponding expert $E_i$. The overall output $y$ is synthesized by aggregating the experts' outputs, each modulated by its respective gating score:
\begin{equation}
    y = \sum_{i=1}^n s_i \cdot (B_i \cdot A \cdot x)
    \tag{7}
\end{equation}
This approach allows for efficient parameterization and specialization of the model's capacity, leveraging the shared matrix $A$ for common transformations and the individual matrices $B_i$ for task-specific adjustments.

\end{enumerate}

\end{document}